\documentclass{article}

\usepackage{amsfonts}
\usepackage{amssymb}
\usepackage{latexsym}
\usepackage{graphicx}

\title{A Genetic Algorithm to Optimize a Tweet for Retweetability}
\author{Ronald Hochreiter \and Christoph Waldhauser}
\date{May 2013}

\begin{document}

\maketitle

\begin{abstract}
Twitter is a popular microblogging
  platform. When users send out messages, other users have the
  abilitiy to forward these messages to their own subgraph. Most
  research focuses on increasing retweetability from a node's
  perspective. Here, we center on improving message style to increase
  the chance of a message being forwarded. To this end, we simulate an
  artifical Twitter-like network with nodes deciding deterministically
  on retweeting a message or not. A genetic algorithm is used to
  optimize message composition, so that the reach of a message is
  increased. When analyzing the algorithm's runtime behavior across a
  set of different node types, we find that the algorithm consistently
  succeeds in significantly improving the retweetability of a
  message.
\end{abstract}

\noindent {\bf Keywords:} Twitter, social network, message style, genetic algorithm, deterministic optimization.

\section{Introduction}
Twitter is a popular microblogging platform, that has been frequently
at the focal point of research. Of special interest has been the complex
network structure that characterizes Twitter networks and the
specifics that govern the propagation of information within Twitter
networks. But how can Twitter users style their messages, so that they
reach furthest?

In this paper we aim at making use of that research by building a
simulation framework to enable researchers to investigate more closely
the propagation of information on Twitter. The simulation framework
is being put to the test by tasking a genetic algorithm with composing
a tweet that reaches furthest in different metrics. In that, we differ
from \cite{kempe2003maximizing} seminal contribution by optimizing
message contents instead of optimizing target audience. The latter
approach in only of limited use in the online scenario, as Twitter
authors cannot influence who will follow them.

This paper is structured as follows. First relevant research regarding
Twitter's networking structure and information diffusion is being
reviewed. We then introduce the simulation framework and describe the
algorithm that was used to obtain optimal tweets. Finally we present
the results and offer some conclusions.


\section{Message diffusion in Twitter networks}
When communicating an actor seeks to get her message across
\cite{shannon2002mathematical}. A central aspect of this process is to
ensure that a message is not only received by the original audience,
but also that this audience spreads that message further on their own
accounts \cite{mcnair2011introduction}. This process has been
researched rather thoroughly from very different aspects: medical
epidemiology \cite{gruhl2004information} and system
dynamics \cite{goldenberg2001talk} to name but a few approaches fielded
to tackle this complex problem. While findings and insights differ, a
common denominator is that message recipients will resend a message if
it passes a recipient's filter, i.e. is to her liking
\cite{rogers2010diffusion}. These filters are domain specific but the
common principle of message diffusion remains true for very diverse
domains.

The advent of micro-blogging has greatly simplified access to message
diffusion data. By looking at e.g.\ Twitter data, connection structure
as well as message contents and meta data are readily available in a
machine readable format. This has produced a wealth of studies
relating to message diffusion on Twitter. In the following, we will
survey recent contributions to the field to distill key factors that
influence message diffusion on Twitter.

In Twitter, users post short messages that are publicly viewable
online and that get pushed to other users following the original
author. It is common practice to cite (retweet) messages of other
users and thus spread them within ones own part of the
network. Messages can contain up to 140 characters including free
text, URL hyperlinks and marked identifiers (hashtags) that show that
a tweet relates to a certain topic. Metadata associated with each
tweet is the point of origin, i.e. the user that posted the tweet, the
time it was posted and the user agent or interface used to post it. On
top of that, the tweets relation to other tweets is available. For
each user, additional meta data is available like the age of the
account, the number of followers, a description and the location.

Twitter networks are typical for the networks of human
communication. They are more complex (i.e. structured and scale-free)
than randomly linked networks with certain users functioning as hubs
with many more connections than would be expected under uniform or
normal distributions. It is useful to think of Twitter networks as
directed graphs with nodes being Twitter users and the following of a
user being mapped to the edges \cite{kwak2010what}. A tweet then
travels from the original author to all directly connected nodes. If
one of the nodes chooses to retweet the message, it is propagated
further down the network.

For average users, Twitter networks' degree distribution follows a
power law and \cite{kwak2010what,java2007we} report the distribution's
exponent to be 2.3 and 2.4 respectively, therefore well within the
range of typical human communication networks. However, there are
extremely popular Twitter authors (celebrities, mass media sites) that
have many more followers than would be expected even under a power-law
distribution.

A distinguishing feature of Twitter is its small-world property. Most
users are connected to any other user using only a small number of
nodes in between. See \cite{kwak2010what} for an overview of Twitter's
small world properties. Despite their findings that for popular users,
the power-law distribution is being violated and average path lengths
between users being shorter than expected, they underscore that
homophily (i.e. similar users are more likely to be in contact) can be
indeed observed and that geographically close users are more likely to
be connected.

Following the notion of message filtering introduced above, it is
clear that Twitter users are selecting messages for propagating them
further according to specific preferences. Applying these preferences
for filtering purposes, they can make use of the message contents
available as listed above. Besides the number of URLs
\cite{yang2010predicting,suh2010want} and hashtags \cite{suh2010want} 
contained, also the free text contents are of importance. According to
\cite{stieglitz2012political,cha2010measuring}, a key aspect in
filtering free text is the polarity and the emotionality of the
message. \cite{stieglitz2012political} also point to the length of
tweet being an important predictor for its retweetability.

Beside message specific filtering criteria, also author specific
filtering can occur. For instance, a Twitter user that has a past
record of being retweeted often, will be more likely to be retweeted
in the future \cite{yang2010predicting,zaman2010predicting}. However,
when styling a single tweet for maximum retweetability, factors like
past popularity or even number of followers \cite{suh2010want} cannot
be influenced and are therefore not represented in the model used.

Shifting the focus from the message recipient to the message sender,
spreading a message as far as possible is a key goal. The success of a
message can be measured using different metrics. In their seminal
work, \cite{yang2010predicting} list three possibilities: One is the
(average) speed a tweet achieves in traversing a network. Another
popularity metric is the scale, that is total number of retweets
achieved. Finally, range can be considered a popularity metric as
well. Here range is the number of edges it takes to travel from the
original author to the furthest retweeter.

In this section we reviewed the latest research related to message
diffusion on Twitter. Key factors influencing the probability of a
tweet being retweeted are the polarity and emotionality of a tweet,
its number of included hyperlinks and hashtags as well as the time of
day it was originally posted. There are other factors influencing
retweet probability, however they are beyond the control of a message
sender and therefore do not apply to the problem at hand. In the next
section we will introduce a simulation framework that can be used to
establish a Twitter-like network to analyze the diffusion principles
of messages governing them.


\section{Simulation framework}

This paper uses the concept of message filtering to simulate the
diffusion of messages in networks and Twitter serves as an example for
this. As detailed above, Twitter users are considered nodes, their
following relationships edges in the network graph. Messages they send
travel from node to node along the graph's edges. The topographical
features of this network, i.e. the distribution of edges, follow the
specifics of scale-free, small-world networks as described above. The
nodes have individual preferences that govern if a message is being
passed on or ignored. In the following we will describe the simulator
used to simulate this kind of network.

Twitter networks exhibit a number of characteristics that we discussed
above. The simulator uses these properties to generate an artificial
network that very similar to Twitter networks. To this end, the number
of connections a node has is drawn from a power-law distribution. In
accordance with the findings reported above, the distribution's
exponent is fixed $2.4$. From these figures, an adjacency matrix is
constructed. As Twitter's following relations are not required to be
reciprocal, the resulting graph is directed. As Twitter contains many
isolated nodes, the resulting graph based on a Twitter-like power-law
distribution also contains a number of isolated nodes. However, these
nodes are irrelevant for the problem at hand, and are thus removed.

Every node is then initialized with a set of uniform random message
passing preferences. The dimensions and their domains are given in
Table~\ref{tab:prefs}. 


\begin{table}[h] 
  \begin{center}
    \caption{Message and node preferences.} \label{tab:prefs}
    \begin{tabular}{ll}
      \hline
      Parameter & Domain\\
      \hline\hline
      Polarity & $-1;1$\\
      Emotionality & $-1;1$\\
      Length & $1;140$\\
      Time & morning, afternoon, night\\
      \# URLs & $0;10$\\
      \# Hashtags & $0;10$\\
      \hline
    \end{tabular}
  \end{center}
 \end{table}

When a message is sent out from the original authoring node, it is
passed on initially to all first-degree followers of that node. Each
follower is then evaluated, if she will pass on the message or
not. This process is repeated until all nodes that have received the
message have been evaluated.

A node's decision on passing the message or not is based on the
preferences of that node. In this model, this decision is purely
deterministic. A node computes the differences between its own preferences and the properties of the message in all six dimensions. If the mean of this differences is lower than some threshold value $\epsilon$, the message is being forwarded. Otherwise, it is discarded.

The simulation framework described above was used to generate an
artificial Twitter-like network for use in this simulation study. To
focus on the principles of message propagation, only a small network
with initially 250 nodes was generated. After removing isolated nodes,
245 nodes with at least 1 connection remained. The average path length
of that network was $4.3$. The maximum of first degree connections
was observed to be at 170 nodes. This is much larger than median and
mean observed to be at $2$ and $3.5$, respectively.

In this section we described how an artificial Twitter-like network
was built using a power-law distribution. This network was paired with
node preferences with respect to the passing on of messages. Using a
deterministic function, each node uses its own preferences and a
message's properties to decide on whether to pass it on or not. In the
following we will describe a genetic algorithm that was used to craft
a message that will reach a maximum number of nodes within that
network.


\section{Algorithm}
In the simulated network, nodes pass on any message they encounter
according to the message properties and their own preferences
regarding these properties. If a sender now wants to maximize the
effect a message has, i.e. to maximize the retweets a tweet will
experience, she has to write a message that meets the expectations of
the right, i.e. highly connected nodes. While topical choices are
obviously important as well, also the right choices regarding message
style influence the probability of a message being retweeted. In this
section we present a genetic algorithm that styles messages so that a
maximum number of nodes retweet it.

The algorithm's chromosome are the message properties as described in
Table \ref{tab:prefs}. An initial population of size $50$ was
initialized with random chromosomes. Using the standard genetic
operators of mutations and crossover, the algorithm was tasked to
maximize the number of nodes that received the message. In the terms
introduced above, this relates to the scale of a message spreading.

To ensure that successful solutions are carried over from one
generation to the next, the top 3 solutions were cloned into the next
generation. This approach of elitism was shown by
\cite{bhandari1996genetic} to positively impact a genetic algorithm's
runtime behavior. Ten percent of every generation was reseeded with
random values to ensure enough fresh material in the gene pool. The
remaining 85 percent of a generation was created by crossing over the
chromosomes of two solutions. To identify solutions eligible for
reproduction, tournament selection using a tournament size of 5 was
implemented. Children's genes were mutated at random. The probability
of a child being mutated was set to be at $0.05$.

In this section we described a genetic algorithm that can maximize the
retweetability of a tweet. Using state of the art genetic operators
and selection mechanisms, a message is being styled so that it will
reach a maximum number of nodes. In the following we describe the
success the algorithm had in fulfilling its task using sender nodes
with a high, medium and low number of first-degree connections.

\section{Results}
The genetic algorithm as described above was used to find optimal
message composition with respect to retweetability for three different
sender nodes. The sender nodes differed in the number of first degree
connections they had. The genetic algorithm described above was
allowed to search for an optimum for 250 generations. Each
optimization run was replicated 50 times with random start values. The
reported result are averages and standard errors across those 50
replications. 

To evaluate the algorithm's performance, two factors are key: the
number of generations it takes to arrive at markedly more successful
messages and the stability of the discovered solutions. While the
former is important to gauge the algorithm's runtime behavior and
suitability for real-world deployment, the latter can reveal insights
on how easy findings can be generalized across different networks. In
the following, these the results relating to these two factors across
all three node types are being described.

\begin{figure}[h]
\begin{center}
\includegraphics[scale=0.35]{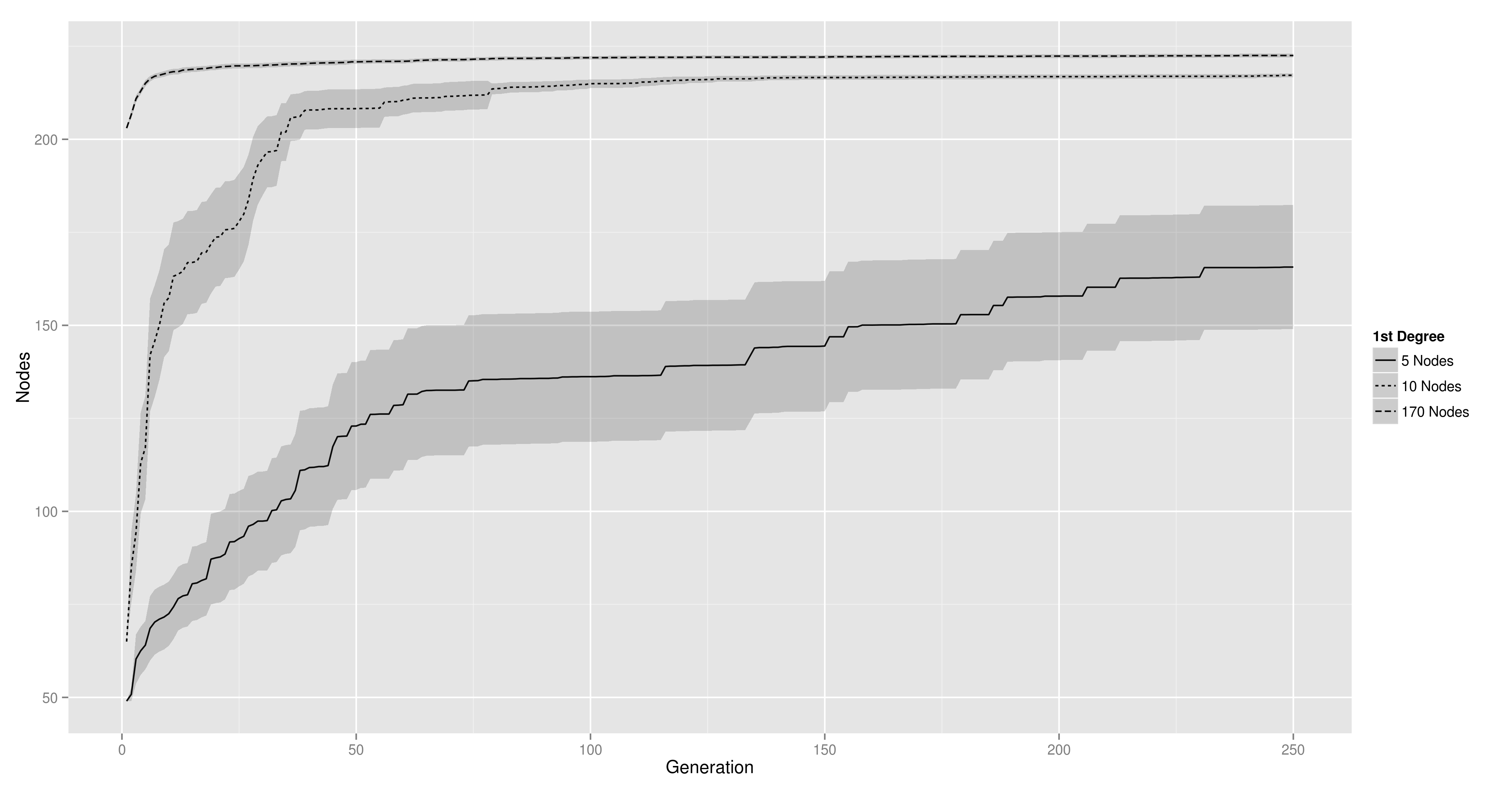}
\caption{Mean fitness as improving over generations for four different kinds of sender nodes. Shaded area is a 95\% confidence interval derived from replicating the optimization 50 times.}\label{fig:fitplot}
\end{center}
\end{figure}

Irrespective of the number of first degree connections a node has,
optimization quickly succeeds in improving the initially random message
styles. Figure \ref{fig:fitplot} depicts the clearly visible trend.

Turning the attention towards stability, the last generation's best solution should be similar across all 50 replications. Table \ref{tab:stab} gives the means and their standard errors for all three kinds of nodes.

\begin{table}[h] 
  \begin{center}
    \caption{Solution stability. Mean and Standard Error (in parentheses).} \label{tab:stab}
    \begin{tabular}{llll}
      \hline
      Parameter & 5 nodes & 10 nodes & 170 nodes\\
      \hline\hline
      Polarity & -0.08 (0.14) & 0.38 (0.17) & 0.32 (0.15)\\
      Emotionality & -0.01 (0.17) & -0.02 (0.22) & 0.06 (0.23)\\
      Length & 82.26 (15.43) & 104.28 (9.80) & 72.18 (14.59)\\
      Time & 2.00 (0.00) & 2.00 (0.00) & 1.60 (0.49)\\
      \# URLs & 5.60 (1.01) & 5.76 (0.98) & 4.00 (0.97)\\
      \# Hashtags & 3.14 (1.05) & 6.06 (0.98) & 6.84 (0.71)\\
      \hline
    \end{tabular}
  \end{center}
 \end{table}

We will discuss these results in the next section and offer some concluding remarks.

\section{Discussion}
The evaluation results provided in the previous section exhibit a
number of peculiarities. Most striking is perhaps, that for high and
medium first degree connections, the algorithm quickly improves the
styling of a message so that many more nodes will retweet it. While
the highly connected sender node has only little room to improve, a
sender with a modest number of connections benefits greatly from
applying the algorithm. For both node types, the algorithm almost
reaches its optimum after 50 generations. For a node with very few
first-degree connections, the optimization process takes much longer
to complete. However, in a subsequent simulation, the algorithm
eventually reached into similar heights, given enough generations. On
average, with only five connected nodes to start from, the algorithm
required 720 generations to reach an optimum beyond 200 nodes.

Another interesting observation is, that the variance in the obtained
solutions increases steadily as the number of first-degree connection
decreases.

In many applications of genetic algorithms, the stability of
identified optimal solutions across replications is a decisive
factor. For the problem at hand, stability is of lesser
importance. When styling a message, apparently different methods lead
to nearly equal performance of the message. 

\section{Conclusion}

In this paper we introduced a genetic algorithm to optimize the
retweetability of tweets. To do this, we simulated a Twitter-like
network and associated each node with a set of preferences regarding
message retweeting behavior. The node's decision is purely
deterministic, based on message properties coming close to a node's
own preferences. The genetic algorithm succeeded in styling messages
so that they became retweeted more widely. Dependent on the number
first-degree connections of the sender node, the fitness of the
algorithm's terminal solution varied. 

This contribution is but a first step in an endeavor to understand the
precise mechanics of message propagation on Twitter. Previous work was
focused on sender node properties. By taking message properties in
account when assessing retweetability, we not only ventured into
uncharted territory, we also discovered new insights regarding the
feasibility of message optimization. 

Our model has a number of limitations, that need to be resolved in
future research. Most prominently, this is the deterministic decision
function. While reasonable for a first model, it would be naive to
assume that nodes' retweet behavior is purely mechanistic. Rather, it
is plausible that the decision to retweet is being influenced to no
small part by chance. Therefore, a stochastic decision function would
be required. We are confident, however, that the genetic algorithm
presented can also optimize a stochastic problem.

Another extension is the calibration of the used Twitter network with
real-life empirical data. This would allow to initialize the nodes not
with uniform random values, but rather with empirically observed ones.


\bibliography{lit}

\begin{thebibliography}{10}

\bibitem{bhandari1996genetic}
D.~Bhandari, C.~A. Murthy, and S.~K. Pal.
\newblock Genetic algorithm with elitist model and its convergence.
\newblock {\em International Journal of Pattern Recognition and Artificial
  Intelligence}, 10(6):731--747, 1996.

\bibitem{cha2010measuring}
M.~Cha, H.~Haddadi, F.~Benevenuto, and P.~K. Gummadi.
\newblock Measuring user influence in {T}witter: The million follower fallacy.
\newblock In {\em Proceedings of the Fourth International Conference on Weblogs
  and Social Media, ICWSM 2010}. The AAAI Press, 2010.

\bibitem{goldenberg2001talk}
J.~Goldenberg, B.~Libai, and E.~Muller.
\newblock Talk of the network: A complex systems look at the underlying process
  of word-of-mouth.
\newblock {\em Marketing Letters}, 12(3):211--223, 2001.

\bibitem{gruhl2004information}
D.~Gruhl, R.~Guha, D.~Liben-Nowell, and A.~Tomkins.
\newblock Information diffusion through blogspace.
\newblock In {\em Proceedings of the 13th International Conference on World
  Wide Web, WWW 2004}, pages 491--501. ACM, 2004.

\bibitem{java2007we}
A.~Java, X.~Song, T.~Finin, and B.~L. Tseng.
\newblock Why we twitter: An analysis of a microblogging community.
\newblock In {\em Advances in Web Mining and Web Usage Analysis, 9th
  International Workshop on Knowledge Discovery on the Web, WebKDD 2007},
  volume 5439 of {\em Lecture Notes in Computer Science}, pages 118--138.
  Springer, 2009.

\bibitem{kempe2003maximizing}
D.~Kempe, J.~M. Kleinberg, and {\'E}.~Tardos.
\newblock Maximizing the spread of influence through a social network.
\newblock In {\em Proceedings of the Ninth ACM SIGKDD International Conference
  on Knowledge Discovery and Data Mining, KDD 2003}, pages 137--146. ACM, 2003.

\bibitem{kwak2010what}
H.~Kwak, C.~Lee, H.~Park, and S.~B. Moon.
\newblock What is {T}witter, a social network or a news media?
\newblock In {\em Proceedings of the 19th International Conference on World
  Wide Web, WWW 2010}, pages 591--600. ACM, 2010.

\bibitem{mcnair2011introduction}
B.~McNair.
\newblock {\em An Introduction to Political Communication}.
\newblock Routledge, 2011.

\bibitem{rogers2010diffusion}
E.~M. Rogers.
\newblock {\em Diffusion of Innovations}.
\newblock Free Press, 2010.

\bibitem{shannon2002mathematical}
C.~Shannon and W.~Weaver.
\newblock {\em The Mathematical Theory of Communication}.
\newblock University of Illinois Press, 2002.

\bibitem{stieglitz2012political}
S.~Stieglitz and L.~Dang-Xuan.
\newblock Political communication and influence through microblogging-an
  empirical analysis of sentiment in {T}witter messages and retweet behavior.
\newblock In {\em 45th Hawaii International International Conference on Systems
  Science (HICSS-45 2012)}, pages 3500--3509. IEEE Computer Society, 2012.

\bibitem{suh2010want}
B.~Suh, L.~Hong, P.~Pirolli, and E.~H. Chi.
\newblock Want to be retweeted? {L}arge scale analytics on factors impacting
  retweet in {T}witter network.
\newblock In {\em Proceedings of the 2010 IEEE Second International Conference
  on Social Computing, SocialCom / IEEE International Conference on Privacy,
  Security, Risk and Trust, PASSAT 2010}, pages 177--184. IEEE Computer
  Society, 2010.

\bibitem{yang2010predicting}
J.~Yang and S.~Counts.
\newblock Predicting the speed, scale, and range of information diffusion in
  {T}witter.
\newblock In {\em Proceedings of the Fourth International Conference on Weblogs
  and Social Media, ICWSM 2010}. The AAAI Press, 2010.

\bibitem{zaman2010predicting}
T.~R. Zaman, R.~Herbrich, J.~Van Gael, and D.~Stern.
\newblock Predicting information spreading in {T}witter.
\newblock In {\em Workshop on Computational Social Science and the Wisdom of
  Crowds, NIPS 2010}, 2010.

\end{thebibliography}
\bibliographystyle{plain}

\end{document}